\documentclass[]{memtensor}
\usepackage{enumitem}
\usepackage[utf8]{inputenc}
\usepackage{url}
\usepackage{graphicx}
\usepackage{booktabs}
\usepackage{amsfonts}
\usepackage{nicefrac}
\usepackage{makecell}
\usepackage{microtype}
\usepackage{amsmath}
\usepackage{etoolbox}
\usepackage{lipsum}
\usepackage{minitoc}
\usepackage{tablefootnote}
\usepackage{threeparttable}
\usepackage{wrapfig}
\usepackage{appendix}
\usepackage{multirow}
\usepackage{ulem}
\useunder{\uline}{\ul}{}
\usepackage{colortbl}
\usepackage{adjustbox}
\usepackage{array}
\usepackage{tabularx}
\usepackage{siunitx}
\usepackage{placeins}
\usepackage{float}
\usepackage{xcolor}
\usepackage{mdframed}
\usepackage{listings}

\renewcommand{\arraystretch}{1.12}
\graphicspath{{img/}}
\IfFileExists{fontawesome5.sty}{
    \usepackage{fontawesome5}
}{
    
    \newcommand{\faGithub}{[GH]}
    \newcommand{\faPlug}{[API]}
    
}

\newcommand{\iconbox}[1]{\makebox[1.5em][c]{\large #1}}

\newcommand{\hficon}{\raisebox{-0.15em}{\IfFileExists{logo/huggingface.png}{\includegraphics[height=1.1em]{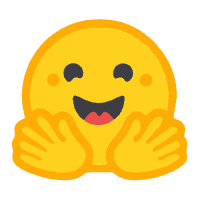}}{[HF]}}}

\newcommand{\figplaceholder}[1]{%
  \fbox{%
    \parbox[c][0.23\textheight][c]{0.9\linewidth}{\centering #1}%
  }%
}

\newcommand{\smartincludegraphics}[2]{%
  \IfFileExists{img/#1}{\includegraphics[width=#2]{#1}}{\figplaceholder{Missing figure: img/#1}}%
}

\lstdefinestyle{promptlisting}{
  basicstyle=\ttfamily\scriptsize,
  breaklines=true,
  breakatwhitespace=false,
  columns=fullflexible,
  keepspaces=true,
  showstringspaces=false,
  upquote=true,
  frame=none
}

\mdfdefinestyle{promptboxstyle}{
  linecolor=gray!60,
  linewidth=0.8pt,
  backgroundcolor=gray!8,
  roundcorner=2pt,
  innertopmargin=8pt,
  innerbottommargin=6pt,
  innerleftmargin=8pt,
  innerrightmargin=8pt,
  skipabove=8pt,
  skipbelow=8pt,
  frametitleaboveskip=0.5em,
  frametitlebelowskip=0.5em,
  frametitlerule=true,
  frametitlerulewidth=0.8pt,
  frametitlebackgroundcolor=gray!20,
  frametitlefont=\bfseries\small
}

\newenvironment{promptbox}[1]{
  \begin{mdframed}[style=promptboxstyle,frametitle={#1}]
}{
  \end{mdframed}
}

\title{%
  \titlefont%
  \parbox{\textwidth}{%
    \centering
    \raisebox{-0.3em}{\includegraphics[height=1.3em]{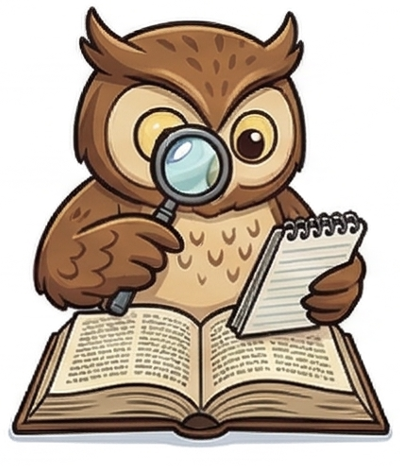}}\hspace{0.5em}%
    {\bfseries MemReader:} From Passive to Active Extraction for Long-Term Agent Memory%
  }%
}

\author[1,*]{Jingyi Kang}
\author[1,*]{Chunyu Li}
\author[2]{Ding Chen}
\author[1]{Bo Tang}
\author[1]{Feiyu Xiong}
\author[1]{Zhiyu Li}

\affiliation[1]{MemTensor (Shanghai) Technology} 
\affiliation[2]{China Telecom Research Institute}
\affiliation[*]{Equal Contribution}

\abstract{
Long-term memory is fundamental for personalized and autonomous agents, yet populating it remains a bottleneck. Existing systems treat memory extraction as a one-shot, passive transcription from context to structured entries, which struggles with noisy dialogue, missing references, and cross-turn dependencies, leading to memory pollution, low-value writes, and inconsistency. In this paper, we introduce the MemReader family for active long-term memory extraction in agent systems: MemReader-0.6B, a compact and cost-efficient passive extractor distilled for accurate and schema-consistent structured outputs, and MemReader-4B, an active extractor optimized with Group Relative Policy Optimization (GRPO) to make memory writing decisions. Under a ReAct-style paradigm, MemReader-4B explicitly evaluates information value, reference ambiguity, and completeness before acting, and can selectively write memories, defer incomplete inputs, retrieve historical context, or discard irrelevant chatter. Experiments on LOCOMO, LongMemEval, and HaluMem show that MemReader consistently outperforms existing extraction-based baselines. In particular, MemReader-4B achieves state-of-the-art performance on tasks involving knowledge updating, temporal reasoning, and hallucination reduction. These results suggest that effective agent memory requires not merely extracting more information, but performing reasoning-driven and selective memory extraction to build low-noise and dynamically evolving long-term memory. Furthermore, MemReader has been integrated into MemOS and is being deployed in real-world applications. To support future research and adoption, we release the models and provide public API access.

\vspace{1.5em}
\noindent
\begingroup
\renewcommand{\arraystretch}{1.2}
\begin{tabular}{@{}l@{}}
\iconbox{\faEnvelope} \textbf{Corresponding Author:} Zhiyu Li (lizy@memtensor.cn)\\
\iconbox{\faGithub} \textbf{Github:} \href{https://github.com/MemTensor/MemOS}{github.com/MemTensor/MemOS} \\
\iconbox{\hficon} \textbf{MemReader-4B:} \href{https://huggingface.co/IAAR-Shanghai/MemReader-4B-thinking}{huggingface.co/IAAR-Shanghai/MemReader-4B-thinking} \\
\iconbox{\faPlug} \textbf{API Documentation:} \href{https://docs.openmem.net/cn/api_docs/core/extract_memory}{docs.openmem.net/api} \\
\iconbox{\faPlug} \textbf{API MemReader-0.6B:} \href{https://docs.openmem.net/cn/self_developed_model/extraction_usage_example}{memos-dashboard.openmem.net/cn/models/}
\end{tabular}
\endgroup
}

\begin{document}
\maketitle

\newpage

\section{Introduction}
Long-term memory is widely viewed as a foundational capability for personalized and persistent agents~\cite{ai_memory_survey,packer2023memgpt}. A usable memory system must not only store history, but also convert ongoing dialogue and documents into representations that are \textbf{retrievable, updatable, and reliable for reasoning}. This conversion process, broadly referred to as memory extraction, hinges on a critical operation memory extraction. In practice, memory extraction is the entry point: it determines what gets written, how it is represented, and whether it can be reused later.

\begin{figure*}[htp]
    \centering
    \smartincludegraphics{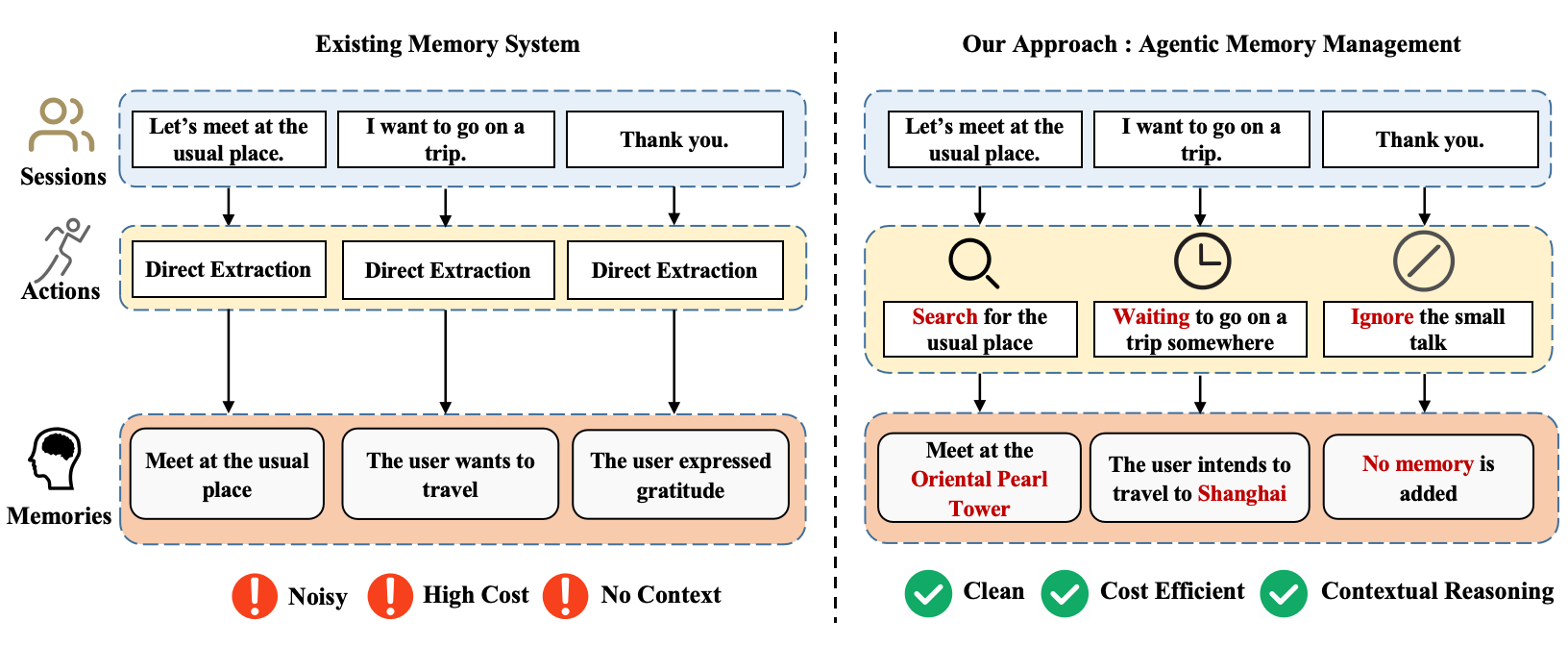}{0.98\textwidth}
    \caption{\textbf{From passive extraction to active memory management.} Conceptual comparison between conventional memory writing pipelines and MemReader. Existing systems mainly translate the current input directly into memory entries, while MemReader explicitly reasons about value, ambiguity, completeness, and update requirements before deciding whether to add, search, buffer, or ignore.}
    \label{fig:overview}
\end{figure*}

Although research on long-term memory has advanced quickly in recent years, most existing systems such as Mem0~\cite{mem0}, Zep~\cite{zep}, and MemOS~\cite{memos} still rely on a relatively passive design~\cite{wang2025text2mem}. In these systems, given the current dialogue or document segment, a large language model directly generates structured memory entries.This treats memory extraction as a one-shot passive extraction task, but in real interaction settings that assumption often fails. Long-term memory is not a static record of the input, rather, it is a dynamic mechanism for maintaining the user state over time. Accordingly, memory extraction is supposed to answer not only ``what should be remembered,'' but also ``should it be remembered at all,'' ``can it be remembered now,'' ``does it need historical disambiguation,'' and ``should new information overwrite old memory.'' Figure~\ref{fig:overview} summarizes this mismatch. Conventional systems conflate semantic extraction with memory management, whereas our method treats memory writing as an explicit state-maintenance problem.

This creates a practical bottleneck. First, most previous systems/pipelines lack value judgment, so low-value information (e.g., small talk) is repeatedly stored and pollute memory. Second, they handle incomplete information poorly, as pronouns, ellipsis, and incremental clarification often require historical context to be properly resolved. Third, they weakly support update and multi-turn fusion, even though user state is frequently revised over time~\cite{zhao2026insideout}. Finally, relying on large general APIs for every extraction step can be costly in deployment.

We argue that the root cause is that existing methods model memory extraction as passive extraction rather than active decision-making. A more reasonable memory module should first judge the value of incoming information, then check whether it is complete or ambiguous, determine whether historical retrieval is required, and finally decide whether to write, buffer, ignore, or update the memory. In other words, long-term memory systems do not merely need ``a better JSON writer'', they need a memory manager that can operate on memory state.

Based on this view, we reframe memory extraction from passive extraction into \textbf{active memory management}. Under this paradigm, we propose MemReader, a family of memory extraction models for long-term agents. MemReader contains two complementary designs. MemReader-0.6B targets cost-sensitive scenarios and distills high-quality structured extraction into a small model, showing that a compact model can still outperform general baselines when the task definition and supervision are sufficiently clear. MemReader-4B introduces a ReAct paradigm, making memory extraction an explicit ``think--act--observe'' process. The model first reasons about whether the current information has long-term value, whether it contains references or cross-turn dependencies, and whether it is complete; then it calls one of several tools, including writing to long-term memory, searching historical memory, buffering incomplete content, or ignoring low-value information. This makes memory extraction and maintenance much closer to real interaction dynamics.

To support this design, we also build a training pipeline tailored for ReAct-style memory extraction. For MemReader-4B, we construct trajectory data covering multiple decision paths such as direct write, retrieval-based disambiguation, buffering, and ignoring. The model is then optimized through reinforcement learning to jointly improve action selection, output quality, and inference efficiency. For MemReader-0.6B, we use bilingual conversation and document memory extraction data to distill the structured extraction ability into a compact model.

We evaluate MemReader on three public benchmarks: LOCOMO, LongMemEval, and HaluMem-Medium. The results show that MemReader-0.6B outperform a GPT-4o-mini-based passive extraction baseline in several settings, validating the effectiveness of the lightweight route. MemReader-4B is especially strong on knowledge update, temporal reasoning, and end-to-end memory usability, demonstrating that explicit decision-making and tool use can reduce noise accumulation, state conflicts, and unusable memory entries.

The main contributions of this paper are as follows:
\begin{itemize}
    \item We propose MemReader, which redefines long-term memory extraction as active memory management. The core goal is to build a low-noise, updatable, and retrievable user-state representation that addresses the value judgment, ambiguity resolution, and state maintenance limitations of existing systems.
    \item We design two complementary models: MemReader-0.6B for a cost-effective small-model route, and MemReader-4B, which explicitly models value judgment, ambiguity resolution, buffering, and updating through a ReAct paradigm.
    \item We validate the method on multiple public benchmarks. The results show clear gains for MemReader, especially in knowledge update, temporal reasoning, and end-to-end memory usability, supporting the active-memory-management direction.
\end{itemize}

\section{Related Work}

Long-term memory and memory extraction are central to personalized and persistent agents~\cite{ai_memory_survey}. Existing work can be grouped into three lines: \textbf{external-memory agent systems}, \textbf{LLM-based structured extraction}, and \textbf{reasoning/tool-use agent paradigms}. Our work lies at the intersection of these lines.

\subsection{Long-Term Memory Systems and External Memory Mechanisms}

As LLMs are deployed in dialogue and assistant systems, many works extend context windows with explicit memory modules. Representative systems include MemGPT, MemoryBank, Mem0, Zep, and MemOS~\cite{packer2023memgpt,memorybank,mem0,zep,memos}. These methods write key interaction information into external stores and retrieve it in later tasks, improving long-horizon QA, preference modeling, and cross-session continuity.

Although these systems have pushed long-term memory forward, most of them still treat memory writing as a relatively simple preprocessing step: extract candidate information from the current interaction and store it directly. This is systemically straightforward, but it makes memory look more like a static cache than a dynamic component that can maintain user state, handle ambiguity, and perform updates. Our focus is precisely this missing piece: we believe the bottleneck of long-term memory systems is not only ``how to retrieve,'' but also ``how to form and maintain memory.''

\subsection{LLM-Based Memory Extraction and Structured Generation}

Another line of work treats memory writing as structured generation. LLMs are used to convert unstructured text into structured events, preferences, constraints, or factual records. In memory systems, this is often implemented by prompting or lightly fine-tuning models to output JSON-like memory entries. This direction is also related to retrieval-augmented generation pipelines~\cite{rag,selfrag}.

The main advantage of this approach is simplicity and portability, and it leverages the semantic understanding already present in general LLMs. However, memory extraction is not exactly the same as standard structured information extraction. It is not meant to cover all input content; instead, it emphasizes selective compression based on future usefulness. As a result, memory extraction naturally includes value judgment, state updating, and cross-turn fusion~\cite{liu2026mememo}. If this is treated as a one-shot structured output task, the model often preserves redundant details, misses the need for later completion of incomplete information, or produces vague expressions when handling pronouns, ellipsis, and historical references. MemReader-0.6B is closest to this route, but we further show that when the training objective matches the task closely enough, even a small model can outperform general baselines on structured memory extraction.

\subsection{Reasoning-Augmented and Tool-Use Agents}

ReAct, Toolformer, Reflexion, and later tool-augmented agent works show that separating ``reasoning'' from ``acting'' can improve robustness and interpretability on complex tasks~\cite{react,toolformer,reflexion}. In settings requiring retrieval, state tracking, multi-step decisions, or delayed judgment, an explicit think--act--observe loop is often more effective than one-shot generation. These methods have shown strong performance in planning, coding, retrieval, and interactive agents~\cite{wang2023voyager,park2023generative}.

We borrow this line of thinking, but our focus is not general task solving; instead, we study \textbf{decision-making in the memory extraction process}. Unlike traditional ReAct work, which mainly targets external tool use, we specialize the tools into the core operations of a memory system: write (\texttt{add}), retrieval-based disambiguation (\texttt{search}), buffering (\texttt{buffer}), and ignoring (\texttt{ignore}). With this design, memory extraction is no longer just ``translate the current input into memory,'' but a dynamic process for maintaining memory state. The experiments also show that this formulation is especially helpful for knowledge update, temporal reasoning, and multi-session consistency.

\subsection{Where This Work Fits}

In summary, existing long-term memory systems have shown the value of external memory plus retrieval augmentation; structured generation methods have demonstrated the feasibility of extracting memory entries from unstructured input; and ReAct-style approaches provide an explicit framework for decision-making and tool use. Our work lies at the intersection of these three directions. We care about practical memory writing, we emphasize structured memory representation, and we further introduce explicit decision-making and tool calls to move memory extraction from passive extraction toward active management. Compared with prior work, the main difference of MemReader is not a larger base model, but a redefinition of the task itself: the goal is not to maximize coverage of the current input, but to maintain a low-noise, updatable, retrievable user-state representation for future interactions.

\section{Method}
\label{sec:problem}

Following the ReAct interaction paradigm~\cite{react}, we formulate memory extraction as a sequential memory-management problem rather than a one-shot structured generation task. At the $t$-th turn, the model observes the current user utterance $x_t$, the long-term memory state $\mathcal{M}_{t-1}$, and the temporary buffer state $\mathcal{B}_{t-1}$. We define the decision state as
\begin{equation}
    s_t = (x_t, \mathcal{M}_{t-1}, \mathcal{B}_{t-1}).
\end{equation}
Given $s_t$, MemReader generates a ReAct-style trajectory
\begin{equation}
    \tau_t = \{(z_t^{(k)}, a_t^{(k)}, o_t^{(k)})\}_{k=1}^{K_t}, \qquad a_t^{(k)} \in \mathcal{A},
\end{equation}
where $z_t^{(k)}$ denotes the internal reasoning trace at step $k$, $a_t^{(k)}$ is the selected tool action, and $o_t^{(k)}$ is the resulting observation. The action space is
\begin{equation}
    \mathcal{A} = \{\texttt{add\_memory},\ \texttt{buffer\_memory},\ \texttt{search\_memory},\ \texttt{ignore\_memory}\}.
\end{equation}
The trajectory updates the memory state through the transition operator
\begin{equation}
    (\mathcal{M}_t, \mathcal{B}_t) = \mathcal{T}(\mathcal{M}_{t-1}, \mathcal{B}_{t-1}, \tau_t),
\end{equation}
where \texttt{add\_memory} writes or updates structured entries in $\mathcal{M}$; \texttt{buffer\_memory} stores incomplete but potentially valuable hypotheses in $\mathcal{B}$; \texttt{search\_memory} retrieves supporting evidence without directly modifying the memory state; and \texttt{ignore\_memory} leaves the state unchanged.

We optimize the policy $\pi_\theta(\cdot\mid s_t)$ to maximize the expected long-term utility of memory extraction:
\begin{equation}
    J(\theta) = \mathbb{E}_{\tau \sim \pi_\theta}\left[\sum_{t=1}^{T} \gamma^{t-1} R(s_t, \tau_t)\right],
\end{equation}
where $R(s_t, \tau_t)$ rewards correct action selection, high-quality memory content, and efficient reasoning, and $\gamma$ is the discount factor.

\subsection{Supervised Fine-Tuning for Warm Start}

We initialize the policy using supervised trajectories. Let $\mathcal{D}_{\mathrm{SFT}}$ denote the supervised dataset, where $y$ is the target token sequence that includes the expected \texttt{<think>} trace, tool call, and structured arguments. We optimize the standard next-token maximum-likelihood objective:
\begin{equation}
    \mathcal{L}_{\mathrm{SFT}}(\theta) = -\mathbb{E}_{(s,y)\sim\mathcal{D}_{\mathrm{SFT}}}\left[\sum_{\ell=1}^{|y|} \log \pi_\theta\big(y_{\ell} \mid s, y_{<\ell}\big)\right].
\end{equation}

This stage teaches the model to follow the output protocol, including properly formatted reasoning traces, well-formed JSON tool calls, and the basic semantics of the four memory actions.

\subsection{Reward Design for Memory Management}

After SFT warm-start, we further optimize with reinforcement learning.
In Agentic RL, long ReAct trajectories introduce three practical bottlenecks. First, training-time optimization over very long token sequences is computationally expensive and unstable. Second, inference-time context growth leads to early-step forgetting, where information from initial reasoning steps is gradually lost. Third, credit assignment across intermediate actions becomes difficult when only sparse terminal outcomes are observed, making it unclear which steps contributed to success or failure.
Our reward design explicitly targets these issues with multi-dimensional shaping, especially the credit-assignment problem.

\textbf{Why Credit Assignment Is Hard.}
With long trajectories, a single terminal reward is too coarse: intermediate steps contribute unequally, while many errors are only exposed at the end. This makes it difficult for policy updates to identify which action should receive positive or negative credit.

\textbf{Multi-Level Reward Shaping.}
We use multi-level shaping to provide denser supervision, combining format validity, action alignment, semantic content quality, and efficiency. This gives actionable feedback at step-level, terminal-decision level, and sequence level.

\textbf{Argument-Level Credit Sinking.}
We further sink credit to the argument level: reward is not only tied to whether an action is called, but also to whether its content is correct, complete, and non-hallucinated.

Given a sampled trajectory $\tau_t$ for dialogue state $s_t$, with token-sequence realization denoted by $y_t$, we define the trajectory-level reward $R(s_t,\tau_t)$, abbreviated as $R_t$, as the weighted sum of four components:
\begin{equation}
R(s_t,\tau_t) \equiv R_t
= \lambda_{\mathrm{fmt}} r_t^{\mathrm{fmt}}
+ \lambda_{\mathrm{align}} r_t^{\mathrm{align}}
+ \lambda_{\mathrm{judge}} r_t^{\mathrm{judge}}
+ \lambda_{\mathrm{eff}} r_t^{\mathrm{eff}},
\label{eq:reward_total}
\end{equation}
where $r_t^{\mathrm{fmt}}$ is the format reward, $r_t^{\mathrm{align}}$ is the action-alignment reward, $r_t^{\mathrm{judge}}$ is the content-quality reward, and $r_t^{\mathrm{eff}}$ is the efficiency reward.
where $r_t^{\mathrm{fmt}}$ is the format reward, $r_t^{\mathrm{align}}$ is the Action-Align reward, $r_t^{\mathrm{judge}}$ is the content-quality reward, and $r_t^{\mathrm{eff}}$ is the efficiency reward.
Here we switch to trajectory tokenization notation: $y_t$ denotes the token sequence generated under state $s_t$.

\paragraph{Format Reward.}
Since MemReader-4B is trained under a ReAct-style protocol, the model output must follow a strict structure with explicit reasoning and action tags, such as \texttt{<think>} and \texttt{<tool\_call>}. We therefore introduce a format reward to encourage outputs that comply with the predefined response schema.
Formally, let
\begin{equation}
    r_t^{\mathrm{fmt}} = \mathbb{I}\big[y_t \in \mathcal{Y}_{\mathrm{valid}}\big],
\end{equation}
where $\mathcal{Y}_{\mathrm{valid}}$ denotes the set of outputs that satisfy all formatting constraints, including:
(1) required tags are present and correctly closed,
(2) the tool-call block is structurally complete,
and (3) the tool arguments can be parsed successfully.
Thus, $r_t^{\mathrm{fmt}}=1$ if the output follows the required format, and $0$ otherwise.
This reward ensures that the model does not merely produce semantically plausible responses, but does so in a form that can be executed reliably by the memory system.

\paragraph{Action-Align Reward.}
To address step-level credit assignment in long trajectories, we introduce a hierarchical action reward with three parts: turn-level alignment, final-decision correctness, and action-distribution consistency.
Let $\mathbf{a}^{\star}=(a_1^{\star},\dots,a_{n}^{\star})$ be the ground-truth action sequence, and $\hat{\mathbf{a}}=(\hat{a}_1,\dots,\hat{a}_{m})$ be the predicted actions.
We define
\begin{equation}
    r_t^{\mathrm{align}} = w_{\mathrm{turn}}\,\bar{r}_{\mathrm{turn}} + w_{\mathrm{final}}\,r_{\mathrm{final}} + w_{\mathrm{dist}}\,r_{\mathrm{dist}},
\end{equation}
where $w_{\mathrm{final}}$ is set to contribute about half of Action-Align reward, reflecting that the terminal decision has the largest business impact.

For turn-level scoring, we compare only the overlapping prefix with length $\ell=\min(n,m)$:
\begin{equation}
    \bar{r}_{\mathrm{turn}} = \frac{1}{\ell}\sum_{i=1}^{\ell} \phi\big(a_i^{\star},\hat{a}_i\big),
\end{equation}
where $\phi(\cdot,\cdot)$ is a severity-aware shaping function: exact matches receive positive reward, while mismatches are penalized by error severity (e.g., wrong \texttt{add\_memory} is penalized more than an unresolved \texttt{search\_memory}). This soft structural alignment avoids over-penalizing near-miss trajectories under sparse supervision.

For final decision quality, we assign stronger terminal reward/penalty based on the final predicted action $\hat{a}_{m}$ and final ground-truth action $a_n^{\star}$.
Incorrect terminal \texttt{add\_memory} decisions receive the strongest penalty because they directly cause memory pollution; correct final decisions receive the largest bonus.

For action-distribution consistency, we compare action-count statistics between trajectories:
\begin{equation}
    r_{\mathrm{dist}} = -\eta_{\mathrm{add}}\,\big|N_{\mathrm{add}}^{\star}-\hat{N}_{\mathrm{add}}\big| - \eta_{\mathrm{search}}\,\big|N_{\mathrm{search}}^{\star}-\hat{N}_{\mathrm{search}}\big|,
\end{equation}
where $N_{\mathrm{add}}$ and $N_{\mathrm{search}}$ denote action counts. Intuitively, this term measures whether the predicted action distribution deviates too much from the ground-truth distribution; large deviations receive additional penalty. It reduces \emph{phantom credit}, where a model matches a few steps but still follows an incorrect overall behavior pattern.

\paragraph{LLM-Judge Reward.}
To evaluate the quality of the extracted memory itself, we employ an LLM-as-a-judge scoring function.
Instead of relying only on exact matching, we ask a judge model to assess the extracted memory content from three complementary aspects:
\textbf{correctness}, \textbf{completeness}, and \textbf{hallucination avoidance}.
Specifically, given the extracted memory $m_t$ and the reference memory annotation $m_t^\star$, the judge model outputs three scalar scores:
\begin{equation}
    s_t^{\mathrm{cor}},\; s_t^{\mathrm{comp}},\; s_t^{\mathrm{hall}} \in [0,1],
\end{equation}
where $s_t^{\mathrm{cor}}$ measures whether the extracted memory is factually consistent with the dialogue context,
$s_t^{\mathrm{comp}}$ measures whether important memory information is sufficiently covered,
and $s_t^{\mathrm{hall}}$ measures the degree to which the output avoids unsupported or fabricated content.
We then define the judge reward as
\begin{equation}
    r_t^{\mathrm{judge}}
    =
    \alpha_{\mathrm{cor}} s_t^{\mathrm{cor}}
    +
    \alpha_{\mathrm{comp}} s_t^{\mathrm{comp}}
    +
    \alpha_{\mathrm{hall}} s_t^{\mathrm{hall}},
\end{equation}
where $\alpha_{\mathrm{cor}}$, $\alpha_{\mathrm{comp}}$, and $\alpha_{\mathrm{hall}}$ are weighting coefficients satisfying.This reward is only applied when the trajectory produces an \texttt{add\_memory} payload.
\begin{equation}
    \alpha_{\mathrm{cor}} + \alpha_{\mathrm{comp}} + \alpha_{\mathrm{hall}} = 1.
\end{equation}
This design allows the reward to capture memory quality at a semantic level:
the model is encouraged not only to extract correct information, but also to retain essential details while suppressing hallucinated content.
Operationally, this component acts as content-level credit assignment: the reward is attached not only to whether \texttt{add\_memory} was called, but also to whether its arguments are correct, complete, and non-hallucinated.

\paragraph{Efficiency Reward.}
A practical memory extractor should compress useful information rather than copy the original dialogue verbatim.
To encourage concise memory extraction, we impose a maximum output-length budget and define an efficiency reward based on the generated length.
Let $L_t$ denote the character length of the model output at turn $t$, and let $L_{\max}$ be the predefined maximum allowed length.
We define
\begin{equation}
    r_t^{\mathrm{eff}} =
    \begin{cases}
        1 - \dfrac{L_t}{L_{\max}}, & L_t \le L_{\max}, \\[6pt]
        -\delta, & L_t > L_{\max},
    \end{cases}
\end{equation}
where $\delta > 0$ is a penalty constant for overly long outputs.
This reward encourages the model to summarize and compress the dialogue into compact memory representations, rather than directly reproducing large spans of the original input.
Together with the action-distribution term in Action-Align reward, it also discourages unnecessary action chains and mitigates context-length growth during long-horizon execution.

\paragraph{Final Reward.}
Combining the components above, the final reward used in GRPO is defined in Eq.~\eqref{eq:reward_total},
where $\lambda_{\mathrm{fmt}}$, $\lambda_{\mathrm{align}}$, $\lambda_{\mathrm{judge}}$, and $\lambda_{\mathrm{eff}}$ are hyperparameters.
In practice, the format reward ensures executable outputs, Action-Align reward provides step-level and terminal-level action credit assignment, the LLM-judge reward improves semantic content quality, and the efficiency reward discourages redundant trajectories.
Together, these four components align optimization with active memory management under long-horizon interaction.

\subsection{GRPO Objective}

Starting from the SFT model, we optimize the model with GRPO\cite{deepseekmath}, a PPO-style method that avoids training a separate critic~\cite{ppo}. For each state $s$, we sample a group of $G$ candidate trajectories $\{y_i\}_{i=1}^{G}$ from the old policy $\pi_{\theta_{\mathrm{old}}}(\cdot \mid s)$ and compute their rewards $\{R_i\}_{i=1}^{G}$. The group-relative normalized advantage is
\begin{equation}
    \hat{A}_i = \frac{R_i - \bar{R}}{\sqrt{\frac{1}{G}\sum_{j=1}^{G}(R_j-\bar{R})^2} + \epsilon},
    \qquad
    \bar{R} = \frac{1}{G}\sum_{j=1}^{G}R_j.
\end{equation}
where $\epsilon > 0$ is a small numerical-stability constant used to avoid division by zero.
For each token position $\ell$ in candidate $y_i$, we define the importance ratio as
\begin{equation}
    \rho_{i,\ell}(\theta) = \frac{\pi_\theta(y_{i,\ell}\mid s, y_{i,<\ell})}{\pi_{\theta_{\mathrm{old}}}(y_{i,\ell}\mid s, y_{i,<\ell})}.
\end{equation}
The clipped GRPO objective is given by
\begin{equation}
\mathcal{J}_{\mathrm{GRPO}}(\theta) =
\mathbb{E}_{s,\{y_i\}}\!\Bigg[\frac{1}{G}\sum_{i=1}^{G}\frac{1}{|y_i|}\sum_{\ell=1}^{|y_i|}
\min\!\Big(\rho_{i,\ell}(\theta)\hat{A}_i,\;
\operatorname{clip}(\rho_{i,\ell}(\theta), 1{-}\epsilon_c, 1{+}\epsilon_c)\hat{A}_i\Big)\!\Bigg]
- \beta\,D_{\mathrm{KL}}\!\big(\pi_\theta \| \pi_{\mathrm{ref}}\big),
\end{equation}
where $\epsilon_c$ denotes the clipping range, and $\pi_{\mathrm{ref}}$ denotes the reference policy used for KL regularization.

Training follows a two-stage pipeline. We first learn an SFT-initialized policy by solving
\begin{equation}
    \theta_{\mathrm{SFT}} = \arg\min_{\theta} \mathcal{L}_{\mathrm{SFT}}(\theta).
\end{equation}
Starting from $\theta_{\mathrm{SFT}}$, we further optimize the policy with GRPO:
\begin{equation}
    \theta^{\star} = \arg\max_{\theta} \mathcal{J}_{\mathrm{GRPO}}(\theta).
\end{equation}

This SFT$+$GRPO formulation aligns well with the task structure: SFT teaches the correct action format and core behaviors, while GRPO sharpens relative preferences among competing decision trajectories under the same memory state.

\section{Model Architecture}

\begin{figure*}[t]
    \centering
    \smartincludegraphics{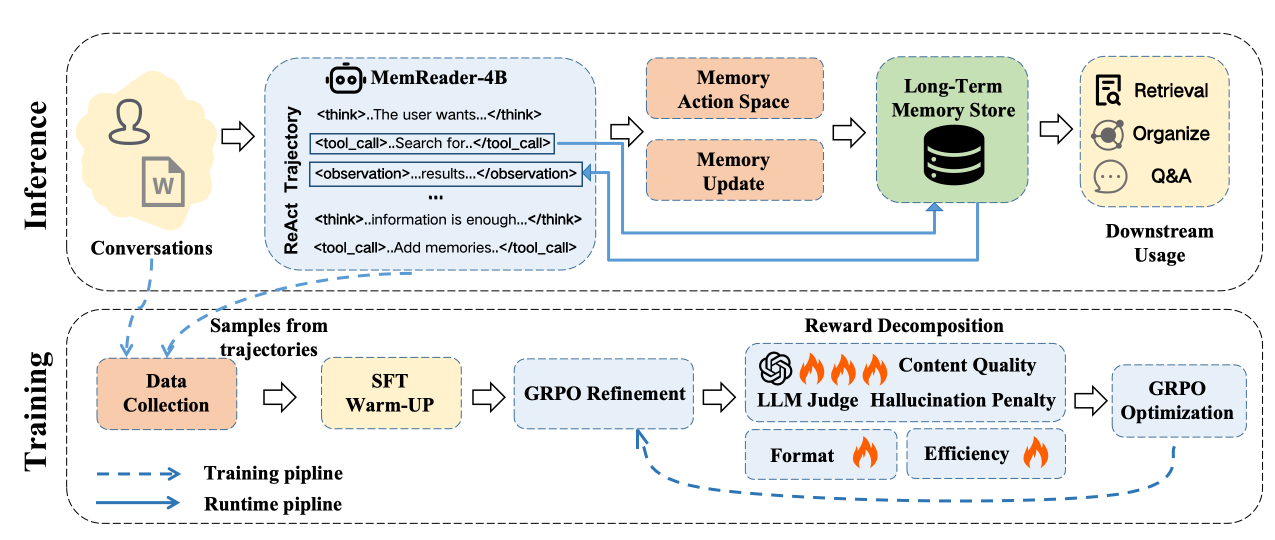}{0.96\textwidth}
    \caption{\textbf{Overall framework of MemReader.} MemReader combines a lightweight structured extractor (MemReader-0.6B) and a reasoning-driven memory manager (MemReader-4B). The 4B model operates in a ReAct loop over memory-specific tools so that memory writing becomes explicit state management rather than passive extraction.}
    \label{fig:framework}
\end{figure*}

Figure~\ref{fig:framework} summarizes the overall design of the MemReader family. The two models share the same goal of forming a low-noise, updatable memory state from dialogue and documents, but they target different deployment regimes. MemReader-4B focuses on active memory management through reasoning and tool use, while MemReader-0.6B focuses on efficient structured extraction for latency- and cost-sensitive settings.

\subsection{MemReader-4B: A ReAct-Based Memory Extraction Model}

MemReader-4B is built on Qwen3-4B~\cite{qwen3} and trained to follow a ReAct (Reasoning + Acting) agent framework\cite{react}. It has both internal thinking (\textit{Think}) and tool-calling (\textit{Action}) abilities. Unlike passive extraction, MemReader-4B first reasons internally about each turn of dialogue and answers three core questions:

\begin{itemize}
    \item \textbf{Q1 Value judgment:} Is the current information worth memorizing? Is it a preference, constraint, important decision, or just low-value chatter or generic knowledge?
    \item \textbf{Q2 Ambiguity detection:} Does the information contain pronouns, ambiguous references, or cross-session links that require retrieval for disambiguation?
    \item \textbf{Q3 Completeness check:} Is the information complete enough to form memory directly, or should it be buffered until later clarification?
\end{itemize}

\paragraph{Tool-calling ability}
Based on the internal reasoning result, MemReader-4B can call four tools. Unlike general-purpose tool-use agents~\cite{toolformer}, these tools are specialized for memory-state operations:
\begin{itemize}
    \item \textbf{\texttt{add\_memory}:} write directly into the long-term memory store when the information is valuable and complete.
    \item \textbf{\texttt{buffer\_memory}:} temporarily store useful but incomplete information until later turns provide more details.
    \item \textbf{\texttt{search\_memory}:} retrieve historical memories to resolve references or ambiguity.
    \item \textbf{\texttt{ignore\_memory}:} discard low-value or generic information.
\end{itemize}

\begin{figure}[t]
    \centering
    \smartincludegraphics{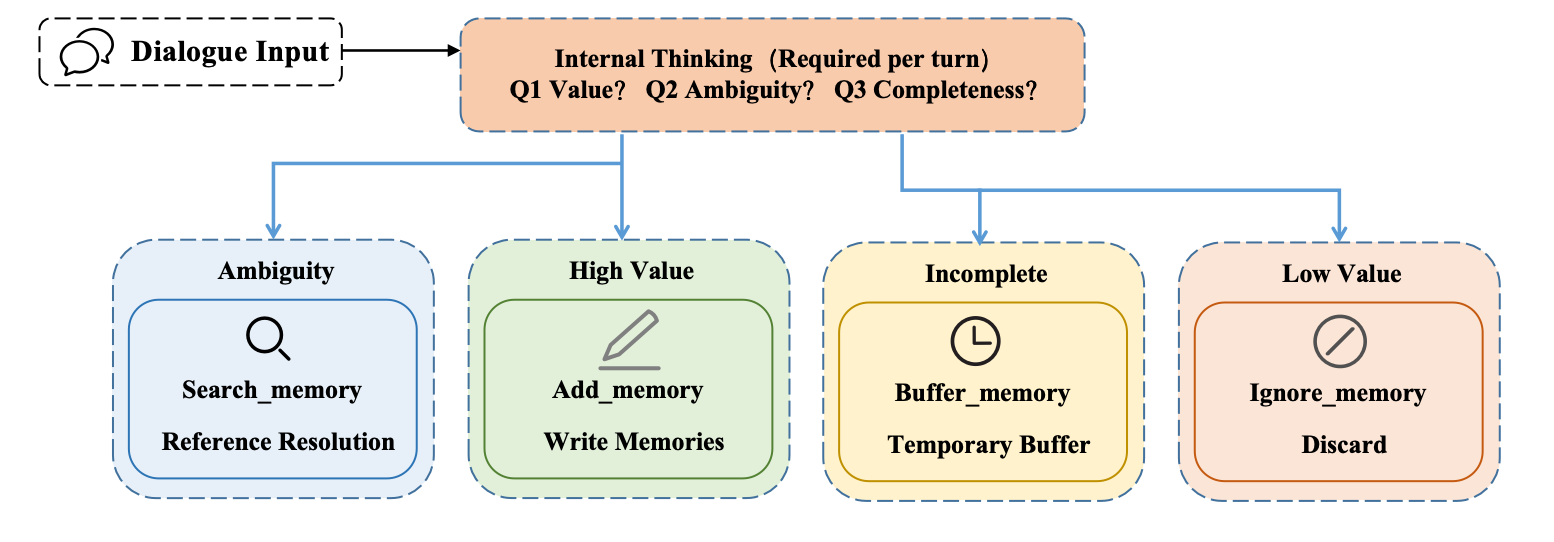}{0.92\linewidth}
    \caption{\textbf{Memory-operation tools used by MemReader-4B.} The tool space is specialized for memory management rather than general task solving, covering direct writing, temporary buffering, retrieval-based disambiguation, and explicit ignoring.}
    \label{fig:tools}
\end{figure}

The tool space is illustrated in Figure~\ref{fig:tools}. The overall pipeline follows a cycle of input $\rightarrow$ think $\rightarrow$ action (tool call) $\rightarrow$ observation (tool result), until the extraction result is finalized or the information is ignored.

\subsection{MemReader-0.6B: A Structured Extraction Model}

MemReader-0.6B is built on Qwen3-0.6B~\cite{qwen3} and trained by distilling high-quality conversation extraction data. It is designed to fit well into the MemOS intermediate workflow, allowing users to combine the 0.6B model with MemOS in practical deployments.

\section{Training}
This section describes the training procedures for both MemReader-4B and MemReader-0.6B. We first detail the data construction pipeline and multi-stage training strategy for MemReader-4B, then briefly describe the distillation-based training of MemReader-0.6B.

\subsection{MemReader-4B Training Process}
MemReader-4B is trained through a dedicated pipeline that combines trajectory-level data construction with multi-stage optimization. We first describe how the training data is constructed to cover diverse memory-management decisions, and then present the three-stage training strategy.

\subsubsection{Training Data Construction}

To train the ReAct capability of MemReader-4B, we designed a dedicated data construction pipeline.

\textbf{Construction idea:} memory extraction requires the model to make different decisions based on value, completeness, and ambiguity at each dialogue turn. We want the training data to cover: (1) complete high-value information that can be written directly (\texttt{add\_memory}); (2) information that must first be retrieved and disambiguated before writing (\texttt{search\_memory} $\rightarrow$ \texttt{add\_memory}); (3) information that is useful but incomplete and should be buffered (\texttt{buffer\_memory}); and (4) low-value information such as small talk or generic knowledge (\texttt{ignore\_memory}). In addition, a single conversation may involve multi-turn buffer-to-add chains, so the data must support cross-turn consistency.

\textbf{Construction method:} we collected diverse multi-turn dialogue datasets covering reference resolution, cross-session association, incomplete information, and low-value information. A strong reasoning model (Gemini-3-Flash-Preview) was used as the teacher model to generate trajectory data with complete Think-Action-Observation chains for each turn. Under the system prompt, the teacher model independently chose one of the four tools and output its internal reasoning through a \texttt{<think>} tag. For scenarios requiring search, we connected a real vector retrieval backend (Milvus) so the teacher could receive realistic observation feedback, ensuring that the search $\rightarrow$ observation $\rightarrow$ add chain was logically correct.

\textbf{Data conversion and chained expansion:} we unified the training data into ShareGPT format (alternating system, human, function\_call, and observation messages) and supported chained multi-turn composition with \texttt{max\_chain\_len=10}. When one turn decided to buffer, the next turn was appended, forming complete chains such as human $\rightarrow$ fc $\rightarrow$ obs $\rightarrow$ human $\rightarrow$ fc $\rightarrow \cdots$ until an add or ignore decision terminated the chain. This allowed one training sample to cover a full buffer-to-add decision path.

\textbf{Quality filtering:} the generated data underwent multiple filtering steps, including JSON validation, tool-call logic checks, and reasoning quality checks. The final dataset contained 7k SFT samples and 3k GRPO samples.

\subsubsection{Multi-Stage Training}

MemReader-4B is trained with a three-stage strategy:
``SFT warm-start $\rightarrow$ DPO alignment attempt $\rightarrow$ GRPO reinforcement learning.'' 
We found this progression important because memory extraction under a ReAct-style paradigm is neither a pure structured prediction problem nor a standard preference learning problem. Instead, it requires the model to jointly satisfy protocol compliance, semantic memory quality, and compact reasoning.

\textbf{Stage 1 --- SFT warm-start:}
We first used the LlamaFactory framework~\cite{zheng2024llamafactory} to fully fine-tune Qwen3-4B on the constructed ReAct trajectories.
The main configuration included a learning rate of 1e-5, cosine decay, 3 epochs, thinking mode enabled (\texttt{enable\_thinking: true}), and \texttt{cutoff\_len=4096}. Training was conducted on 8 $\times$ A800 80GB GPUs with DeepSpeed ZeRO-3.
This stage primarily teaches the model the basic interaction protocol of the task, including producing valid \texttt{<think>} traces, well-formed \texttt{<tool\_call>} outputs, and the core behavior patterns of the four memory actions.
In practice, SFT is necessary because subsequent reinforcement learning becomes much more stable when the model already knows how to follow the required think--act format.

\textbf{Stage 2 --- DPO alignment attempt:}
After SFT, we tried to further align the model with Direct Preference Optimization (DPO)~\cite{dpo}.
However, DPO worked poorly for this task in our experiments.
Although the training loss decreased rapidly, the rewards of both chosen and rejected samples kept dropping, the rejected reward fell below -30, and gradients vanished very early.
Our analysis suggests that memory extraction is not a task with sharply separable preference pairs.
Under the same dialogue state, two candidate outputs may both be partially reasonable, differing only in subtle aspects such as memory completeness, hallucination level, or compression quality.
When the SFT model is already reasonably strong, the gap between positive and negative samples becomes too small and fine-grained for pairwise preference learning to provide a stable optimization signal.
As a result, DPO did not reliably improve the model, and we therefore moved to GRPO.

\textbf{Stage 3 --- GRPO reinforcement learning:}
We then adopted Group Relative Policy Optimization in the verl framework~\cite{sheng2024hybridflow} and conducted multi-turn dialogue training.
Compared with DPO, GRPO is better suited to this task because it compares multiple sampled trajectories under the same state and learns from their relative differences, rather than relying on a single chosen/rejected pair.
This is especially important for memory extraction, where multiple outputs may all be executable yet differ in protocol compliance, semantic faithfulness, and compression efficiency.
In our setting, GRPO provides a more informative and stable learning signal for ranking these subtly different trajectories.

Another practical reason GRPO works better is that memory extraction quality is inherently multi-dimensional.
A good output should not only be correct, but should also follow the required ReAct format, preserve important information, avoid hallucinated content, and remain concise enough to function as memory rather than a raw transcript.
Our reward design directly reflects these requirements.
In particular, we found it important to explicitly constrain the length of the \texttt{<think>} segment.
When the reasoning trace becomes excessively long, the model tends to over-elaborate on latent hypotheses and is more likely to introduce unsupported details, often leading to more severe hallucinations in the final extracted memory.
For this reason, we impose a character-level efficiency constraint and penalize overly long thinking traces.
This encourages the model to reason only as much as needed for memory decisions, rather than extending speculative chains that hurt final memory quality.

More broadly, the reward design is effective because its three components are closely aligned with the practical deployment requirements of the memory system.
The format reward ensures that outputs can be parsed and executed reliably in the ReAct pipeline.
The LLM-judge reward encourages extracted memories to be correct, complete, and free of hallucinations at the semantic level.
The efficiency reward discourages copying the original dialogue and pushes the model toward compact memory representations.
Together, these rewards optimize not only textual plausibility, but also usable memory updates in a real agentic memory system.

\begin{sloppypar}
Unless otherwise noted, we used the following GRPO settings:
\texttt{train\_batch\_size=8}, \texttt{rollout.n=8}, \texttt{lr=1e-6},
\texttt{temperature=0.7}, \texttt{max\_assistant\_turns=16},
\texttt{max\_user\_turns=15}, and \texttt{max\_response\_length=768}. 
\end{sloppypar}

\subsection{MemReader-0.6B Training Process}

MemReader-0.6B uses a distillation-based route to generate structured memory extraction results. The training data covers both conversation memory extraction and document memory extraction with samples in Chinese and English. During training, we apply JSON validation, field-completeness checks, and date-parsing accuracy checks, then use supervised fine-tuning (SFT) to distill the extraction ability into Qwen3-0.6B.

\section{Experiments}

\subsection{Benchmarks}

We evaluate the MemReader series on the following three public benchmarks:
\begin{itemize}
    \item \textbf{LOCOMO}~\cite{locomo}: a long-dialogue memory benchmark with Single Hop, Multi Hop, Temporal, Open Domain, Overall, and F1 metrics.
    \item \textbf{LongMemEval}~\cite{longmemeval}: a benchmark for long-term memory systems with six dimensions: single-session-preference, single-session-assistant, temporal-reasoning, multi-session, knowledge-update, and single-session-user.
    \item \textbf{HaluMem-Medium}~\cite{halumem}: a hallucination-oriented memory benchmark that evaluates retrieval, update, and QA ability, with metrics such as recall, precision, hallucination rate, and omission rate.
\end{itemize}

For all benchmarks, we use GPT-4.1-mini as both the response model and the evaluation model. We also record average token consumption per extraction as an efficiency metric.

Unless otherwise noted, higher values indicate better performance. For token consumption and omission/hallucination-style error rates, lower is better. In all tables, the best value in each metric column is \textbf{bold}, and the second-best value is \underline{underlined}.

\subsection{LOCOMO Results}

Table~\ref{tab:locomo} reports the main LOCOMO results across hop-based reasoning, temporal understanding, open-domain memory QA, and overall score. Both MemReader variants remain competitive with stronger or larger baselines, while exhibiting different trade-offs between efficiency and end-task performance.

MemReader-0.6B shows strong performance on structured extraction quality, achieving the best Temporal score and the highest F1, indicating its effectiveness in producing precise and well-structured memory representations. In contrast, MemReader-4B-GRPO achieves the best Overall score, with clear improvements on Multi-Hop and Open-Domain reasoning, suggesting stronger capability in handling complex reasoning and broader memory retrieval scenarios.This contrast highlights the complementary roles of the two variants: a lightweight extractor optimized for structured accuracy, and a larger model that better captures complex reasoning and memory utilization in end tasks.

\begin{table}[h]
\centering
\caption{LOCOMO main accuracy and average token consumption per extraction. F1 values for some baselines are reported as - when corresponding original papers do not provide directly mappable LOCOMO overall F1 under this setting.}
\small
\begin{adjustbox}{max width=\linewidth}
\begin{tabular}{lccccccc}
\toprule
\textbf{Model} & \textbf{Single Hop} & \textbf{Multi Hop} & \textbf{Temporal} &
\textbf{Open Domain} & \textbf{Overall} & \textbf{F1} & \textbf{Token} \\
\midrule
MemoryOS & 67.30\% & 59.34\% & 42.26\% & 59.03\% & 60.11\% & - & 5500 \\
Mem0 & 68.97\% & 61.70\% & 58.26\% & 50.00\% & 64.20\% & - & \underline{1000} \\
MemU & 74.91\% & 72.34\% & 43.61\% & 54.17\% & 66.67\% & - & 4000 \\
MemOS(4o-mini) & 84.06\% & 73.16\% & \underline{75.90\%} & 57.29\% & 78.70\% & \underline{51.90\%} & 1854 \\
MemReader-0.6B & \underline{84.70\%} & \underline{76.95\%} & \textbf{76.22\%} & 53.40\% & \underline{79.56\%} & \textbf{52.54\%} & 1976 \\
MemReader-4B-SFT & 81.88\% & 76.12\% & 71.02\% & \underline{62.15\%} & 77.33\% & 47.77\% & \textbf{784} \\
MemReader-4B-GRPO & \textbf{85.37\%} & \textbf{81.44\%} & 75.80\% & \textbf{65.62\%} & \textbf{81.42\%} & 49.45\% & 1950 \\
\bottomrule
\end{tabular}
\end{adjustbox}
\label{tab:locomo}
\vspace{0.5em}
\end{table}

\subsection{LongMemEval Results}

Table~\ref{tab:compare} presents a unified comparison of open-source memory systems on LongMemEval.

Overall, MemReader-4B-GRPO achieves the best performance, matching the highest Overall score while also leading in knowledge-update and temporal-reasoning, indicating strong capability in maintaining and evolving long-term memory states. Compared to other systems, it demonstrates clear advantages in handling dynamic and time-sensitive information.MemReader-0.6B also performs competitively, achieving the best performance on multi-session tasks, suggesting that well-structured extraction can remain effective even in scenarios requiring cross-session memory consistency.These results suggest that long-term memory performance depends not only on extraction quality, but also on effective memory updating and temporal reasoning, where larger models with active management strategies provide clear benefits.
\begin{table}[h]
\centering
\caption{Main accuracy and average token consumption per extraction on LongMemEval. SS denotes single-session tasks.}
\small
\setlength{\tabcolsep}{4.5pt}
\begin{adjustbox}{max width=\linewidth}
\begin{tabular}{lcccccccc}
\toprule
\textbf{Method} & \textbf{Token} & \textbf{SS-User} & \textbf{SS-Asst} & \textbf{SS-Pref} & \textbf{Multi-S} & \textbf{Know. Upd} & \textbf{Temp. Reas} & \textbf{Overall} \\
\midrule
MIRIX & - & 72.85\% & 63.63\% & 53.33\% & 30.07\% & 52.56\% & 25.56\% & 43.49\% \\
Zep & 1600 & 92.90\% & \underline{75.00\%} & 53.30\% & 47.40\% & 74.40\% & 54.10\% & 63.80\% \\
Mem0 & 1066 & 82.86\% & 26.78\% & 90.00\% & 63.15\% & 66.67\% & 72.18\% & 66.40\% \\
Memobase & 1541 & 92.85\% & 23.21\% & 80.00\% & 66.91\% & \underline{89.74\%} & 75.93\% & 72.40\% \\
MemU & \textbf{523} & 67.14\% & 19.64\% & 76.67\% & 42.10\% & 41.02\% & 17.29\% & 38.40\% \\
MemOS & 1400 & 95.71\% & 67.86\% & \textbf{96.67\%} & 70.67\% & 74.26\% & 77.44\% & 77.80\% \\
EverMemOS & 2800 & \textbf{97.14\%} & \textbf{85.71\%} & \underline{93.33\%} & \underline{73.68\%} & \underline{89.74\%} & 77.44\% & \textbf{83.00\%} \\
MemReader-0.6B & 1166 & 95.71\% & \underline{75.00\%} & 90.00\% & \textbf{75.18\%} & 82.05\% & 75.90\% & \underline{80.20\%} \\
MemReader-4B-SFT & 963 & \underline{97.10\%} & 69.64\% & 90.00\% & 71.42\% & 85.80\% & \underline{78.19\%} & 80.00\% \\
MemReader-4B-GRPO & \underline{922} & 94.29\% & 73.21\% & 90.00\% & \underline{73.68\%} & \textbf{91.03\%} & \textbf{84.21\%} & \textbf{83.00\%} \\
\bottomrule
\end{tabular}
\end{adjustbox}
\label{tab:compare}
\end{table}

\subsection{HaluMem-Medium Results}
Table~\ref{tab:halumem} reports the full HaluMem-Medium results across memory extraction, updating, and question answering. MemReader-4B-GRPO achieves the highest scores in most extraction metrics, including Recall (96.57\%), Weighted Recall (97.19\%), and F1 (98.21\%), while also obtaining the best update correctness (94.55\%) and the lowest update omission rate (5.12\%). MemReader-0.6B leads in extraction accuracy (95.66\%) and QA omission (12.14\%), showing strong structured output quality even at a smaller scale. In the QA stage, MemOS retains the best correctness and hallucination rate, suggesting that downstream answer generation may also depend on retrieval strategy and response model behavior beyond the extraction module alone. Overall, the MemReader models demonstrate clear advantages in extraction and updating, confirming that higher-quality memory writing propagates benefits through the entire memory pipeline.

\begin{table}[h]
\centering
\caption{
Evaluation results of all memory systems on HaluMem. \textbf{R} denotes Recall, \textbf{Target P} denotes Target Memory
Precision, \textbf{Acc.} denotes Accuracy, \textbf{FMR} denotes False Memory Resistance, \textbf{F1} denotes Memory Extraction
F1-score, \textbf{C} denotes Correct Rate (Accuracy), \textbf{H} denotes Hallucination Rate, and \textbf{O} denotes Omission Rate.  
}
\footnotesize
\setlength{\tabcolsep}{2.6pt}
\renewcommand{\arraystretch}{1.05}

\begin{adjustbox}{max width=\textwidth}
\begin{tabular}{lcccccccccccc}
\toprule
\multirow{2}{*}{\textbf{System}} 
& \multicolumn{6}{c}{\textbf{Memory Extraction}} 
& \multicolumn{3}{c}{\textbf{Memory Updating}} 
& \multicolumn{3}{c}{\textbf{Question Answering}} \\
\cmidrule(lr){2-7} \cmidrule(lr){8-10} \cmidrule(lr){11-13}
& \textbf{R$\uparrow$} 
& \textbf{W-R$\uparrow$} 
& \textbf{T-P$\uparrow$} 
& \textbf{Acc.$\uparrow$} 
& \textbf{FMR$\downarrow$} 
& \textbf{F1$\uparrow$}
& \textbf{C$\uparrow$} 
& \textbf{H$\downarrow$} 
& \textbf{O$\downarrow$}
& \textbf{C$\uparrow$} 
& \textbf{H$\downarrow$} 
& \textbf{O$\downarrow$} \\
\midrule

Zep 
& -- & -- & -- & -- & -- & --
& 47.28\% & 0.42\% & 52.31\%
& 55.47\% & 21.92\% & 22.62\% \\

Mem0-Graph 
& 43.28\% & 65.52\% & 87.20\% & 61.86\% & 55.70\% & 57.85\%
& 24.50\% & \textbf{0.26\%} & 75.24\%
& 54.66\% & 19.28\% & 26.06\% \\

Mem0 
& 42.91\% & 65.03\% & 86.26\% & 60.86\% & 56.80\% & 57.31\%
& 25.50\% & 0.45\% & 74.02\%
& 53.02\% & \underline{19.17\%} & 27.81\% \\

Supermemory 
& 41.53\% & 64.76\% & 90.32\% & 60.83\% & 51.77\% & 56.90\%
& 16.37\% & 1.15\% & 82.47\%
& 54.07\% & 22.24\% & 23.69\% \\

Memobase 
& 14.55\% & 25.88\% & 92.24\% & 32.29\% & 80.78\% & 25.13\%
& 5.20\% & 0.55\% & 94.25\%
& 35.33\% & 29.97\% & 34.71\% \\

MemOS 
& 74.07\% & 84.81\% & 86.25\% & 59.55\% & 44.94\% & 79.70\%
& 62.11\% & 0.42\% & 37.48\%
& \textbf{67.23\%} & \textbf{15.17\%} & 17.59\% \\

MemReader-0.6B 
& 88.40\% & 91.38\% & 99.82\% & \textbf{95.66\%} & 33.84\% & 93.76\%
& 82.69\% & 0.77\% & 16.51\%
& \underline{59.48\%} & 28.38\% & \textbf{12.14\%} \\

MemReader-4B-SFT 
& \underline{93.56\%} & \underline{95.49\%} & \underline{99.86\%} & 91.31\% & \textbf{18.24\%} & \underline{96.61\%}
& \underline{90.78\%} & \textbf{0.26\%} & \underline{8.74\%}
& 56.30\% & 28.44\% & 15.26\% \\

MemReader-4B-GRPO
& \textbf{96.57\%} & \textbf{97.19\%} & \textbf{99.91\%} & \underline{91.98\%} & \underline{19.18\%} & \textbf{98.21\%}
& \textbf{94.55\%} & \underline{0.32\%} & \textbf{5.12\%}
& 56.99\% & 28.67\% & \underline{14.34\%} \\
\bottomrule
\end{tabular}
\end{adjustbox}
\label{tab:halumem}
\end{table}

\FloatBarrier

\subsection{Analysis}

Across all three benchmarks, a consistent pattern emerges: improvements in memory systems are not solely driven by better extraction accuracy, but by the ability to maintain, update, and utilize memory as a dynamic state.

MemReader-0.6B demonstrates that high-quality structured extraction can be achieved with a compact model when the task is well-defined. Its strong performance suggests that memory extraction quality depends not only on model scale, but also on task formulation and representation design, making lightweight extractors a practical and cost-efficient solution.

In contrast, MemReader-4B shows clear advantages in scenarios that require maintaining consistency over time, such as knowledge updating, temporal reasoning, and multi-session interaction. These improvements indicate stronger capability in managing evolving memory states, including resolving ambiguity, updating outdated information, and determining when memory is sufficiently complete.

Results on HaluMem-Medium further suggest that these advantages extend beyond individual components. Improvements in extraction and updating lead to more stable behavior across the memory pipeline, reducing inconsistencies and improving downstream usability.

Taken together, these findings point to a shift in how memory systems should be designed. Rather than treating memory as a static extraction problem, effective long-term systems require explicit mechanisms for maintaining a coherent, low-noise, and updatable user state. In this context, MemReader-4B represents a step toward treating memory as an actively managed process rather than a passive output.

\section{Conclusion}

This paper introduces the MemReader family of memory extraction models and reframes long-term memory construction from passive extraction to active management. MemReader-0.6B shows that task-oriented distillation enables a lightweight model to match or surpass GPT-4o-mini--level extraction quality, while MemReader-4B adopts the ReAct paradigm to model key decisions including writing, retrieval, buffering, and filtering and moves memory processing toward explicit state maintenance. Experiments on LOCOMO, LongMemEval, and HaluMem show clear gains, particularly in knowledge updating, temporal reasoning, and end-to-end memory usability.

More broadly, our results suggest that the core of a long-term memory system is not to extract more information from input, but to build and maintain a low-noise, updatable, and retrievable user-state representation that supports downstream reasoning. Accordingly, memory modules in long-term agents should be treated as first-class components that continuously maintain state, rather than as static extractors.

Several directions remain open: extending the tool set to support memory editing, conflict detection, and hierarchical abstraction; evaluating stability and long-term gains in realistic online interaction settings beyond current benchmarks; and jointly optimizing memory extraction, organization, and response generation. We hope this work provides a clear path toward explicit, controllable, and maintainable memory management in long-term agents.

\bibliographystyle{unsrt}
\bibliography{main}

\appendix
\section{Appendix A. Prompt Template Examples}

This appendix provides practical prompt templates used by MemReader-0.6B and MemReader-4B for memory extraction and memory management.

\subsection{A.1 MemReader-0.6B Conversation Extraction Prompt }

This prompt is designed for high-fidelity structured extraction in multilingual conversations, with strict constraints on temporal normalization, reference resolution, and JSON output validity.

\begin{promptbox}{MemReader-0.6B Prompt Template}
\begin{lstlisting}[style=promptlisting]
SIMPLE_STRUCT_MEM_READER_PROMPT = """You are a memory extraction expert.
Your task is to extract memories from the perspective of user,
based on a conversation between user and assistant.
This means identifying what user would plausibly remember,
including their own experiences, thoughts, plans, or relevant
statements and actions made by others (such as assistant)
that impacted or were acknowledged by user.
Please perform:
1. Identify information that reflects user's experiences,
beliefs, concerns, decisions, plans, or reactions,
including meaningful input from assistant that user
acknowledged or responded to.
If the message is from the user, extract user-relevant memories;
if it is from the assistant, only extract factual memories that
the user acknowledged or responded to.

2. Resolve all time, person, and event references clearly:
   - Convert relative time expressions (e.g., "yesterday,"
     "next Friday") into absolute dates using the message
     timestamp if possible.
   - Clearly distinguish between event time and message time.
   - If uncertainty exists, state it explicitly (e.g., "around June 2025," "exact date unclear").
   - Include specific locations if mentioned.
   - Resolve all pronouns, aliases, and ambiguous references
     into full names or identities.
   - Disambiguate people with the same name if applicable.
3. Always write from a third-person perspective, referring to user as
"The user" or by name if name mentioned, rather than using first-person ("I", "me", "my").
For example, write "The user felt exhausted..." instead of "I felt exhausted...".
4. Do not omit any information that user is likely to remember.
   - Include all key experiences, thoughts, emotional responses, and plans -- even if they seem minor.
   - Prioritize completeness and fidelity over conciseness.
   - Do not generalize or skip details that could be personally meaningful to user.
5. Please avoid any content that violates national laws and
regulations or involves politically sensitive information in
the memories you extract.

Return a single valid JSON object with the following structure:

{
  "memory list": [
    {
      "key": <string, a unique, concise memory title>,
      "memory_type": <string, Either "LongTermMemory" or "UserMemory">,
      "value": <A detailed, self-contained, and unambiguous
                memory statement written in English if the input
                conversation is in English, or in Chinese if the
                conversation is in Chinese>,
      "tags": <A list of relevant thematic keywords
               (e.g., ["deadline", "team", "planning"])>
    },
    ...
  ],
  "summary": <a natural paragraph summarizing the above
              memories from user's perspective, 120-200 words,
              same language as the input>
}

Language rules:
- The `key`, `value`, `tags`, `summary` fields must match
  the mostly used language of the input conversation.
  If the input is Chinese, output in Chinese.
- Keep `memory_type` in English.

Always respond in the same language as the conversation.

Conversation:
${conversation}

Your Output:"""
\end{lstlisting}
\end{promptbox}

\subsection{A.2 MemReader-4B Memory-Management Prompt}

This prompt emphasizes explicit decision-making (add / buffer / ignore / search) and encourages retrieval-based disambiguation before writing uncertain memories.

\begin{promptbox}{MemReader-4B Prompt Template}
\begin{lstlisting}[style=promptlisting]
You are a memory extraction expert. Analyze conversations and decide whether to extract memories.

## Task
Process the dialogue and decide: add (extract), buffer (wait), ignore, or search for context.

## Principles
- Extract if it has long-term value (user or assistant)
- Use third-person ("The user"), never "I" or "me"
- Normalize time to absolute dates using session time
- Resolve all pronouns and ambiguous references

## Preferences
- Prioritize search when encountering ambiguous info ("he", "that thing")
- Only buffer when search cannot resolve AND user hasn't finished speaking
- Do NOT buffer out of laziness

## Memory Format (for add_memory)
{
  "memory_list": [
    {
      "key": "title",
      "memory_type": "LongTermMemory/UserMemory",
      "value": "statement",
      "tags": ["tag"]
    }
  ],
  "summary": "paragraph"
}

## Current State
- Buffer: {buffer_summary}
- Session Time: {session_time}
\end{lstlisting}
\end{promptbox}

\section{Appendix B. Data Construction Details}

This appendix summarizes key data-construction details for MemReader-4B. One major source of trajectory supervision is generated with a stronger teacher model (Gemini-3-Flash), then filtered and converted into ReAct-style training traces.

\subsection{B.1 Teacher Prompt for ReAct Trajectory Construction (Gemini-3-Flash)}

\begin{promptbox}{Gemini-3-Flash System Prompt Template}
\begin{lstlisting}[style=promptlisting]
MEMORY_EXTRACTION_REACT_SYSTEM_PROMPT = """
# You are a Memory Management Expert. You process dialogue through multiple rounds of Thought, Action, and Observation.

# Task: Decide how to handle the current conversation utterance (Extract Memory, Buffer, Ignore, or Search Context).

# Principles:
- Do not distinguish between the user or the assistant; extract if it has long-term value.
- References must be explicit; do not use "I" or "me"; must include specific names or roles.
- Normalize time (combine with current time {session_time}) and facts.

# Preferences:
- Do not be hasty to `add`. If there is potential background information that needs to be supplemented, prioritize `search`.
- When encountering ambiguous information (e.g., "he", "that thing"), you MUST prioritize `search` to try and find the answer in the history.
- Only choose `buffer` when `search` cannot resolve the issue, or when it is obvious the user hasn't finished speaking (a new topic).
- Do not choose `buffer` out of laziness. If you can complete the information via `search` and `add` immediately, that is the highest priority behavior.

# Workflow:
Thought N -> Action N -> Observation N
Eventually end with Action N: finish[...]

# State:
Buffer: {buffer_summary}
Current Time: {session_time}
Current Utterance: {text}

# Output Format (Strictly Adhere):
Thought [N]: ...
Action [N]: ...
Observation [N]: ...

# Action Types:
1. add[]: Information is complete and important; extract immediately.
2. search[query]: Review history. Use this when the dialogue contains pronouns (it, that person), specific references (last time, that project), or implicit background, AND the answer likely exists in past conversation records.
3. buffer[reason]: Wait for the future. Use this when the user starts a new topic but details are not yet unfolded (e.g., "I plan to buy something..."), or the information is brand new but too vague, requiring reliance on the user's future utterances to complete.
4. ignore[reason]: No substantive content / small talk / repetition.
5. finish[action]: End processing and output the final decision (add/buffer/ignore). Only output finish[action]; the [] should contain ONLY the determined action keyword, not the specific content of the action.
"""
\end{lstlisting}
\end{promptbox}

\subsection{B.2 Few-Shot Examples Used for Teacher Trace Generation}

\begin{promptbox}{Gemini-3-Flash Few-Shot Template}
\begin{lstlisting}[style=promptlisting]
FEW_SHOT_EXAMPLES = """
## Example 1: Direct Extraction
Dialogue: User "I just finished my first marathon, 4 hours 30 minutes, felt so accomplished!" (2023-10-15 14:30)
Thought 1: The user shared an important achievement, and the information is complete (event, time, data, emotion), so it can be extracted directly.
Action 1: finish[add]
Observation 1: Ready to extract.

## Example 2: Context Search Required
Dialogue: User "Regarding that plan, I decided to go with Plan B." (2023-10-15 15:00)
Thought 1: "That plan" and "Plan B" are ambiguous references; I need to search the context.
Action 1: search[user's project plan]
Observation 1: Search Result: User previously mentioned a "New Product Plan" containing Option A, Option B (steady progress), and Option C.
Thought 2: Combining with context, the user chose to make steady progress on the new product. The decision is important, and the information is now complete.
Action 2: finish[add]
"""
\end{lstlisting}
\end{promptbox}

\section{Appendix C. Case Studies}

This appendix presents three representative examples from our evaluation on the LOCOMO dataset, illustrating the three key decision outcomes of our ReAct-based memory extraction model: \textbf{Add}, \textbf{Buffer}, and \textbf{Ignore}. Each case includes (1) the current conversational input, (2) the model's ReAct trajectory with search, observation, and reasoning, and (3) the final memory output or decision rationale.It is worth noting that the search return results shown in the example are all from the actual database tool call returns. Here, we use the mivus database.

\subsubsection*{Example A -- Add: Enriching a Memory via Retrieved Context}

\textbf{Input -- Current Conversation Turn}

\begin{promptbox}{Example A Input}
\begin{lstlisting}[style=promptlisting]
Jon [2023-01-20T16:04:00+00:00]: Sorry to hear that! I'm starting a dance studio 'cause I'm passionate about dancing and it'd be great to share it with others.
Gina [2023-01-20T16:04:00+00:00]: That's cool, Jon! What got you into this biz?
\end{lstlisting}
\end{promptbox}

\textbf{Model ReAct Trace}

\begin{promptbox}{Example A ReAct Trace}
\begin{lstlisting}[style=promptlisting]
Thought 1: Jon is starting a dance studio because of a passion for dancing. Gina is asking for the background of this business. The information about Jon's intent to start a dance studio is clear and has long-term value, but Gina's question suggests there might be more depth or history to this decision that hasn't been fully revealed yet. However, the fact that Jon is starting a dance studio is a concrete fact. I should check if there's any prior context about Jon's dancing or business plans to see if this is a continuation or a new development.

Action 1: search_memory
{
  "query": "Jon's dancing or business plans"
}

Observation 1: The search results confirm Jon's background as a banker and his recent job loss, which provides context for his decision to start a dance studio. The current utterance provides the specific nature of the business and his motivation. Although Gina is asking for more background, the fact that Jon is starting a dance studio is a complete and significant update to his status.

Thought 2: I have enough information to extract this.

Action 2: add_memory
{
  "memory_list": [
    {
      "key": "Dance studio startup plan",
      "memory_type": "UserMemory",
      "value": "On January 20, 2023, Jon informed Gina that he is planning to start a dance studio. He explained that the decision is driven by his passion for dancing and his desire to share that passion with other people.",
      "tags": ["business", "dance studio", "career change", "passion"]
    }
  ],
  "summary": "Extracted 1 memories."
}
\end{lstlisting}
\end{promptbox}

\textbf{Output -- Extracted Memory / Final Decision}

\begin{promptbox}{Example A Output}
\begin{lstlisting}[style=promptlisting]
{
  "key": "Dance studio startup plan",
  "memory_type": "UserMemory",
  "value": "On January 20, 2023, Jon informed Gina that he is planning to start a dance studio. He explained that the decision is driven by his passion for dancing and his desire to share that passion with other people.",
  "tags": ["business", "dance studio", "career change", "passion"]
}
\end{lstlisting}
\end{promptbox}

\subsubsection*{Example B -- Buffer: Deferred Extraction for Incomplete Information}

\textbf{Input -- Current Conversation Turn}

\begin{promptbox}{Example B Input}
\begin{lstlisting}[style=promptlisting]
Evan [2023-05-18T13:47:00+00:00]: My old prius broke down, decided to get it repaired and sell it. Glad you asked, we went to Rockies, check it out.
Sam [2023-05-18T13:47:00+00:00]: Wow! Looks amazing. When did you get to go there?
\end{lstlisting}
\end{promptbox}

\textbf{Model ReAct Trace}

\begin{promptbox}{Example B ReAct Trace}
\begin{lstlisting}[style=promptlisting]
Thought 1: Evan provides details about his old Prius (repaired and sold) and the destination of his trip (Rockies). Sam asks for the timing of the trip. While the Prius information is complete, the specific timing of the Rockies trip is pending Evan's answer to Sam's question. I should check if there's any prior mention of the Prius or the trip planning to provide better context.

Action 1: search_memory
{
  "query": "Evan's Prius and Rockies trip"
}

Observation 1: The search for previous context returned nothing, meaning this is new information. Evan has explained why he sold his old Prius (it broke down, repaired, and sold) and where he went (Rockies). However, Sam has just asked when he went.

Thought 2: Since the specific time is currently unknown and likely to be provided in the next utterance, I should buffer this to capture the complete timeline of the trip rather than creating a partial memory.

Action 2: buffer_memory
{
  "reason": "The search for previous context returned nothing, meaning this is new information. Evan has explained why he sold his old Prius (it broke down, repaired, and sold) and where he went (Rockies). However, Sam has just asked when he went. Since the specific time is currently unknown and likely to be provided in the next utterance, I should buffer this to capture the complete timeline of the trip rather than creating a partial memory."
}
\end{lstlisting}
\end{promptbox}

\textbf{Output -- Extracted Memory / Final Decision}

\begin{promptbox}{Example B Output}
\begin{lstlisting}[style=promptlisting]
Decision: buffer
Reason: The search for previous context returned nothing, meaning this is new information. Evan has explained why he sold his old Prius (it broke down, repaired, and sold) and where he went (Rockies). However, Sam has just asked when he went. Since the specific time is currently unknown and likely to be provided in the next utterance, I should buffer this to capture the complete timeline of the trip rather than creating a partial memory.
\end{lstlisting}
\end{promptbox}

\subsubsection*{Example C -- Ignore: Redundant Utterance Already on Record}

\textbf{Input -- Current Conversation Turn}

\begin{promptbox}{Example C Input}
\begin{lstlisting}[style=promptlisting]
Audrey [2023-03-27T13:10:00+00:00]: Cool, gonna give it a try. Thanks for the suggestion!
Andrew [2023-03-27T13:10:00+00:00]: No problem! Let me know how you like it. Have fun hiking!
\end{lstlisting}
\end{promptbox}

\textbf{Model ReAct Trace}

\begin{promptbox}{Example C ReAct Trace}
\begin{lstlisting}[style=promptlisting]
Thought 1: Audrey and Andrew are concluding a conversation about hiking. Audrey expresses intent to try a specific suggestion, and Andrew mentions "hiking". The reference "it" in "give it a try" is ambiguous. I need to search the context to understand what Audrey is referring to before deciding whether to extract or ignore.

Action 1: search_memory
{
  "query": "hiking suggestion Audrey Andrew"
}

Observation 1: The search results clarify that Andrew recommended Fox Hollow as a hiking trail and mentioned spotting birds. Audrey's "it" refers to the Fox Hollow trail or the suggestion to hike there. The current utterance is a polite closing where Audrey accepts the recommendation and Andrew wishes her a fun trip.

Thought 2: This is social pleasantries and doesn't provide new substantive information beyond what was already recorded (the recommendation of Fox Hollow).

Action 2: ignore_memory
{
  "reason": "The search results clarify that Andrew recommended Fox Hollow as a hiking trail and mentioned spotting birds. Audrey's \"it\" refers to the Fox Hollow trail or the suggestion to hike there. The current utterance is a polite closing where Audrey accepts the recommendation and Andrew wishes her a fun trip. This is social pleasantries and doesn't provide new substantive information beyond what was already recorded (the recommendation of Fox Hollow)."
}
\end{lstlisting}
\end{promptbox}

\textbf{Output -- Extracted Memory / Final Decision}

\begin{promptbox}{Example C Output}
\begin{lstlisting}[style=promptlisting]
Decision: ignore
Reason: The search results clarify that Andrew recommended Fox Hollow as a hiking trail and mentioned spotting birds. Audrey's "it" refers to the Fox Hollow trail or the suggestion to hike there. The current utterance is a polite closing where Audrey accepts the recommendation and Andrew wishes her a fun trip. This is social pleasantries and doesn't provide new substantive information beyond what was already recorded (the recommendation of Fox Hollow).
\end{lstlisting}
\end{promptbox}

\end{document}